\documentclass[10pt,twocolumn,letterpaper]{article}

\usepackage{iccv}              
\usepackage{multirow}
\usepackage{placeins}
\usepackage{soul}

\definecolor{iccvblue}{rgb}{0.21,0.49,0.74}
\usepackage[pagebackref,breaklinks,colorlinks,allcolors=iccvblue]{hyperref}

\def\paperID{6} 
\def\confName{ICCV}
\def\confYear{2025}

\title{Adapt, But Don’t Forget: Fine-Tuning and Contrastive Routing for Lane Detection under Distribution Shift}

\author{
Mohammed Abdul Hafeez Khan$^{1}$ \quad Parth Ganeriwala$^{1}$ \quad Sarah M. Lehman$^{2}$ \\ Siddhartha Bhattacharyya$^{1}$ \quad Amy Alvarez$^{1}$ \quad Natasha Neogi$^{2}$\\
$^{1}$Florida Institute of Technology, USA \\
$^{2}$NASA Langley Research Center, USA \\
{\tt\small \{mkhan, pganeriwala2022, alvareza2023\}@my.fit.edu} \quad
{\tt\small sbhattacharyya@fit.edu} \\
{\tt\small \{natasha.a.neogi, sarah.lehman\}@nasa.gov}
}

\begin{document}
\maketitle
\begin{abstract}
Lane detection models are often evaluated in a closed-world setting, where training and testing occur on the same dataset. We observe that, even within the same domain, cross-dataset distribution shifts can cause severe catastrophic forgetting during fine-tuning. To address this, we first train a base model on a source distribution and then adapt it to each new target distribution by creating separate branches, fine-tuning only selected components while keeping the original source branch fixed. Based on a component-wise analysis, we identify effective fine-tuning strategies for target distributions that enable parameter-efficient adaptation. At inference time, we propose using a supervised contrastive learning model to identify the input distribution and dynamically route it to the corresponding branch. Our framework achieves near-optimal F1-scores while using significantly fewer parameters than training separate models for each distribution.
\end{abstract}    
\section{Introduction}
\label{sec:intro}

Deep learning has become the dominant paradigm in computer vision with pretrained models serving as the foundation for state-of-the-art systems across tasks such as image classification~\cite{he2016deep, dosovitskiy2020image, khan2022classification, 11014782}, semantic segmentation~\cite{long2015fully, chen2017deeplab, ronneberger2015u}, and object detection~\cite{chen2017rethinking, he2017mask, khan2024alina, khan2022detection}. While highly effective on in-distribution data, these models often fail under distribution shift, that is, when test data diverge from training distributions~\cite{zhang2019recent, guan2021domain}. This limitation is especially critical in downstream applications like lane detection~\cite{zakaria2023lane}, where safe deployment requires robustness across diverse real-world environments. Recent advances have produced a range of lane detection architectures such as segmentation-based~\cite{pan2018spatial, zheng2021resa, neven2018towards, hou2019learning, ko2021key, lee2017vpgnet, loshchilov2017decoupled}, row-wise~\cite{qin2020ultra, qin2022ultra, han2022laneformer, liu2021condlanenet, ganeriwala2023cross}, anchor-based~\cite{tabelini2021keep, zheng2022clrnet, zheng2025clrnetv2, honda2024clrernet}, and parametric methods~\cite{tabelini2021polylanenet, liu2021end, chen2023bsnet}.  These approaches differ both in how they represent lane lines (e.g., pixel masks, row-wise positions, polynomial curves) and in their trade-offs between instance discrimination, inference, and post-processing complexity.


However, despite the architectural progress and large-scale benchmarks, lane detection models are typically evaluated on a single dataset (e.g., CULane~\cite{pan2018spatial} or CurveLanes~\cite{xu2020curvelane}), reflecting a closed-world assumption that overlooks generalization across datasets and environments~\cite{ganeriwala2023cross}. For instance, a model trained on one-lane dataset may perform poorly on another where lanes originate from different image regions and follow more diverse topologies, thereby violating the model’s learned assumptions. Moreover, when a model trained on car lanes is evaluated on a distribution like airport taxiways (AssistTaxi~\cite{ganeriwala2023assisttaxi}), where visual semantics differ markedly from typical road lanes, we observe that the F1-score drops to near-zero.

The difficulty of generalizing across datasets and distributions is rooted in how lane detection models internalize dataset-specific priors during training. They learn to associate visual cues with patterns that dominate their source distribution, embedding strong assumptions about lane geometry, spatial orientation, and camera perspective. These assumptions are encoded hierarchically across the network: early layers capture low-level textures like road markings while deeper layers and task-specific heads model layout regularities, curvature, and anchor parameter relationships. In anchor-based models, for instance, the network head learns to predict offsets from predefined anchors based on where lane markings typically appear, such as the left and right sides in car-lane datasets. When exposed to a distribution with different curvature, lane density, or semantic context---such as taxiway markings that usually appear in the center of an image---these priors often fail to hold, leading to feature misalignment and degraded scene interpretability. This suggests that the model has not truly learned a distribution-invariant understanding of what constitutes a "lane" in terms of geometry, continuity, and spatial semantics, but has instead internalized anchor patterns and positional constraints tied to the biases of its training data.

While fine-tuning is the most common approach for adapting pretrained models to new data, it often leads to catastrophic forgetting~\cite{kirkpatrick2017overcoming}, that is, a sharp drop in performance on the original training distribution after adaptation. This occurs because the same network parameters that encode useful features for the source distribution are updated to fit the new one, resulting in the overwriting of previously learned representations~\cite{kemker2018measuring, lopez2017gradient}. This phenomenon has been studied extensively in continual learning~\cite{mirzadeh2020understanding, nguyen2020dissecting, ramasesh2020anatomy, kemker2018measuring, lopez2017gradient}. Classical fine-tuning heuristics attempt to reduce forgetting by freezing early layers or limiting updates to task-specific heads, under the assumption that low-level features are more general and stable. Yet, we observe that forgetting persists even under such constraints, suggesting that semantic knowledge is distributed more broadly across the network. This raises a key question: \textit{Can we empirically characterize the roles of different network components in adaptation and forgetting under distribution shift, and use this understanding to inform more targeted and efficient fine-tuning strategies?}

\paragraph{Contributions.} 
In this work, we address the challenge of adapting a lane detection model to a new distribution while preserving its original performance using minimal trainable parameters. To this end, we first conduct a \textit{component-wise analysis} under distribution shift to empirically characterize the \textit{adaptation–retention tradeoff} across different network modules. Based on these insights, we introduce a \textit{modular branching} strategy, where only target-specific components are fine-tuned while the source branch remains fixed—preserving source performance while enabling efficient target adaptation. To support distribution-aware inference, we train a \textit{supervised contrastive model} to classify input distributions and dynamically route them to the appropriate branches. We base our study on \textit{CLRerNet}~\cite{honda2024clrernet}, a state-of-the-art anchor-based lane detection model, and evaluate across three datasets—CULane~\cite{pan2018spatial}, CurveLanes~\cite{xu2020curvelane}, and AssistTaxi~\cite{ganeriwala2023assisttaxi}. Our framework enables \textit{parameter-efficient adaptation} with no source forgetting, and we further validate its generality on ERFNet to examine how \textit{model capacity influences adaptability}.


    


\section{Related Works}
\subsection{Lane Detection and Generalization}

Foundational lane detection models span segmentation-based~\cite{pan2018spatial, zheng2021resa, neven2018towards, hou2019learning, ko2021key, lee2017vpgnet, loshchilov2017decoupled}, row-wise~\cite{qin2020ultra, qin2022ultra, han2022laneformer, liu2021condlanenet}, anchor-based~\cite{tabelini2021keep, zheng2022clrnet, zheng2025clrnetv2, honda2024clrernet}, keypoint-based~\cite{ko2021key, qu2021focus, wang2022keypoint}, and parametric~\cite{tabelini2021polylanenet, liu2021end, chen2023bsnet} paradigms, each embedding distinct assumptions about lane structure. Among these, anchor-based models such as LaneATT~\cite{tabelini2021keep}, CLRNet~\cite{zheng2022clrnet}, and CLRerNet~\cite{honda2024clrernet} have gained traction for their balance of performance and efficiency, leveraging structured anchor templates for direct lane regression. Row-wise approaches like UFLD~\cite{qin2020ultra} and CondLaneNet~\cite{liu2021condlanenet} exploit strong spatial priors to achieve efficient decoding, while parametric methods~\cite{tabelini2021polylanenet} offer compact polynomial representations but often struggle under occlusion or complex lane topologies.

However, robustness to distribution shift remains poorly understood. Architectural biases---such as those embedded in anchor heads or spatial decoders---can hinder adaptability, while compact backbones like ERFNet, though promising for real-time deployment, remain underexplored in this context. Addressing these gaps requires moving beyond end-to-end evaluation toward understanding the role of individual components in enabling generalization.



\subsection{Contrastive Learning and Distribution Adaptation}

Contrastive learning has proven effective for structuring feature space by bringing similar samples closer and pushing dissimilar ones apart~\cite{khosla2020supervised, chen2020simple}. Initially developed for self-supervised representation learning~\cite{he2020momentum, grill2020bootstrap}, it has since been extended to long-tailed classification~\cite{zhu2022balanced}, few-shot detection~\cite{fan2020few}, and semantic segmentation~\cite{wang2021exploring}. In open-set detection, OpenDet~\cite{han2022expanding} leverages contrastive learning to separate known and unknown classes via a Contrastive Feature Learner. In lane detection, recent methods such as LaneCorrect~\cite{nie2025lanecorrect}, DACCA~\cite{zhou2024unsupervised}, and CLLD~\cite{zoljodi2024contrastive} apply contrastive losses for cross-domain alignment or robustness, though they primarily focus on unsupervised settings.

Most domain adaptation approaches in lane detection rely on adversarial learning~\cite{chen2019domain, hu2022sim, tzeng2017adversarial}, self-training~\cite{hoyer2023mic, lian2019constructing}, pseudo-labeling~\cite{li2022multi}, or weak supervision~\cite{das2023weakly, lv2020weakly}, treating the model holistically without examining internal component behavior. While methods like viewpoint normalization~\cite{muthalagu2020lane} and WSDAL~\cite{zhou2024towards} aim to align distributions, they do not provide fine-grained control over adaptation.

\section{Methodology}
\label{sec:method}



\begin{figure*}[t]
    \centering
    \includegraphics[width=0.920\linewidth]{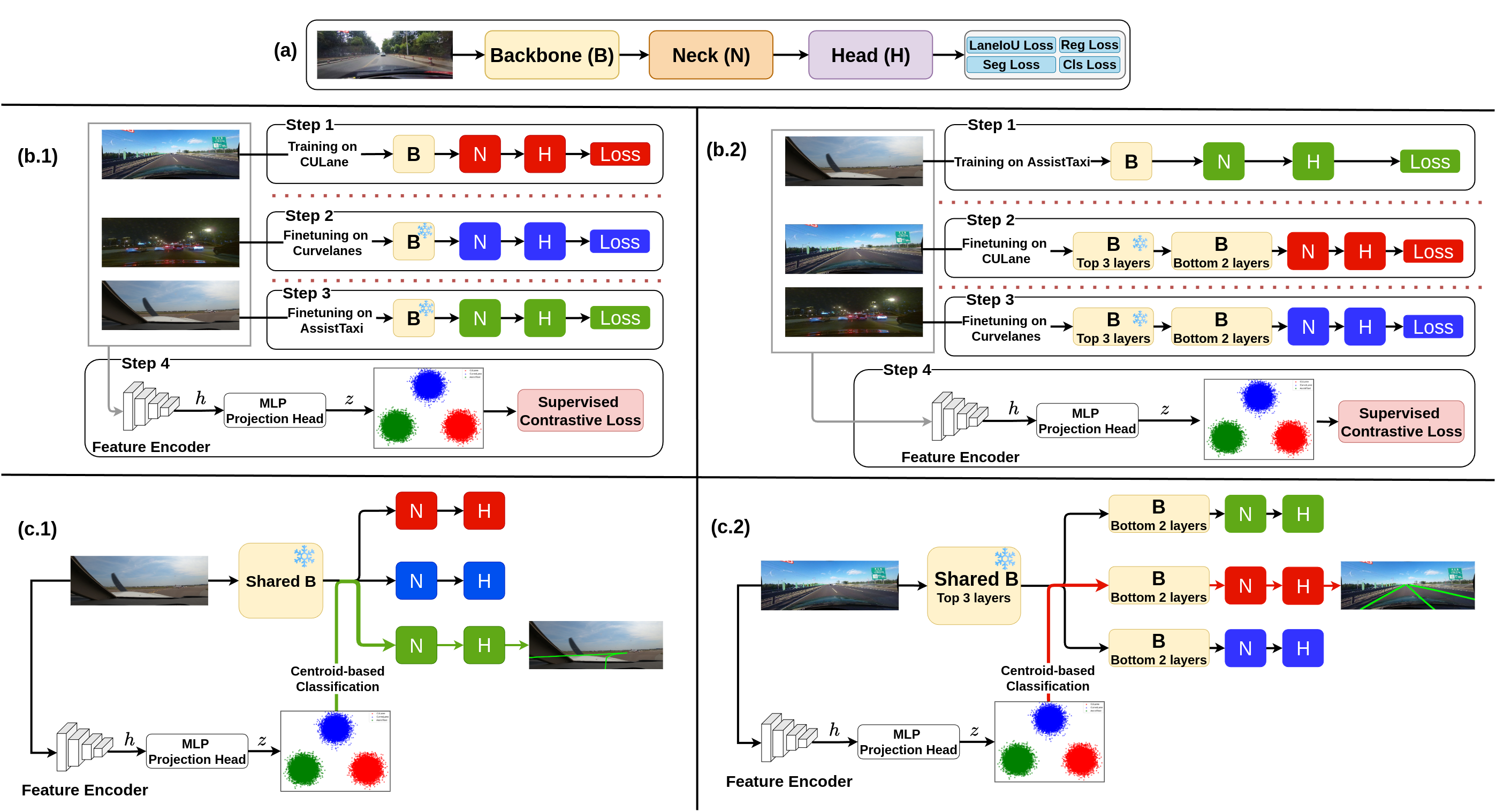}
    \caption{
\textbf{(a)} Overview of CLRerNet.  
\textbf{(b.1)} and \textbf{(c.1)} illustrate one routing configuration. In Step 1, the model is trained on the source distribution (CULane). In Steps 2 and 3, \texttt{B} is frozen, and the \texttt{N} and \texttt{H} modules are \textit{cloned and fine-tuned} for each target distribution. In Step 4, a supervised contrastive learning (SCL) model is trained to classify the input distribution. During inference \textbf{(c.1)}, the SCL model routes shared \texttt{B} features to the corresponding \texttt{N+H} branch. When CULane or CurveLanes is the source, Routing@\texttt{N+H} yields the best results (see Table~\ref{tab:routing}, highlighted in red and blue, respectively).
\textbf{(b.2)} and \textbf{(c.2)} show a deeper routing configuration where AssistTaxi is the source. Here, the top 3 layers of \texttt{B} are frozen. During inference \textbf{(c.2)}, the SCL model classifies the input and routes features from the shared top 3 layers of \texttt{B} to the corresponding branch. Routing@\texttt{B$^{(k=2)}$+N+H} yields the best results for AssistTaxi (see Table~\ref{tab:routing}, highlighted in green).
These configurations show that SCL-based routing can be \textit{dynamically positioned} for efficient, distribution-specific adaptation.
}
    \label{fig:method}
\end{figure*}


\subsection{An Overview of CLRerNet}

CLRerNet is a row-wise anchor-based lane detector that iteratively refines a set of learnable lane priors. Each lane is represented by a tuple \((x_a, y_a, \theta_a, l)\) indicating its starting point, orientation, and length. Let \(P_0 = \{(x_a, y_a, \theta_a, l)\}_n\) denote the initial set of $n$ anchors. The network performs a cascade of refinement stages over feature pyramid levels \(\{L_0, L_1, L_2\}\), enhancing spatial precision from high-level to low-level features. At each stage \(t\), lane-aligned features are sampled along anchor priors \(P_{t-1}\), enhanced via cross-attention with the global feature map, and passed through fully connected layers to predict refinements \((\delta x_a^t, \delta y_a^t, \delta \theta_a^t, \delta l^t)\), classification score \(c^t\), and row-wise offsets \(\delta x_{\text{row}}^t \in \mathbb{R}^{N_{\text{row}}}\). The anchor parameters are updated iteratively as:

\begin{equation}
P_t = P_{t-1} + \Delta P_t, \quad \Delta P_t = (\delta x_a^t, \delta y_a^t, \delta \theta_a^t, \delta l^t)
\label{eq:anchor_update}
\end{equation}

Final lane predictions are decoded from the refined anchor parameters and local offsets. The network is trained with a combined loss:
\begin{equation}
\mathcal{L}_{\text{total}} = \lambda_0 \mathcal{L}_{\text{reg}} + \lambda_1 \mathcal{L}_{\text{cls}} + \lambda_2 \mathcal{L}_{\text{seg}} + \lambda_3 \mathcal{L}_{\text{LaneIoU}}
\label{eq:total_loss}
\end{equation}

where $\mathcal{L}_{\text{reg}}$ is a Smooth-L1 loss on anchor parameters: start point coordinate \((x_a, y_a)\), theta angle \(\theta_a\), and lane length regression \(l\). $\mathcal{L}_{\text{cls}}$ is the focal loss between predictions and labels, $\mathcal{L}_{\text{seg}}$ is an auxiliary
cross-entropy loss for per-pixel segmentation mask, and $\mathcal{L}_{\text{LaneIoU}}$ is the IoU loss
between the predicted lane and ground truth.

The backbone $\mathcal{B}_{\theta_B}$ (ResNet, DLA, or ERFNet) extracts low- and mid-level visual features from the input image. The neck $\mathcal{N}_{\theta_N}$, typically a feature pyramid network (FPN)~\cite{lin2017feature}, fuses multi-scale features to provide context-aware representations. The detection head $\mathcal{H}_{\theta_H}$ samples features along each anchor-aligned lane prior and refines them using a cross-attention mechanism that incorporates contextual information from the global feature map. These features are used to regress anchor parameters, row-wise offsets, and confidence scores. Since the features are sampled along initial anchor locations, the anchor-based architectures assume alignment between the learned priors and the actual lane distributions. However, this alignment may break under distribution shift such as in unseen environments with differing anchor origins or orientations, leading to mislocalized features and reduced prediction quality.

\subsection{Problem Statement}

Let $\mathcal{D}_S = \{(x_i^S, y_i^S)\}_{i=1}^{N_S}$ be a labeled source dataset and $\mathcal{D}_T = \{(x_j^T, y_j^T)\}_{j=1}^{N_T}$ a labeled target dataset from a different distribution (but same domain), where $x$ denotes an input image and $y$ is its corresponding lane annotation encoding pixel-wise lane locations. A lane detection model $f_\theta$, instantiated as CLRerNet with parameters $\theta = \{\theta_B, \theta_N, \theta_H\}$ for backbone, neck, and head respectively, is first trained on $\mathcal{D}_S$. When fine-tuned on $\mathcal{D}_T$, the model exhibits a performance drop on $\mathcal{D}_S$ due to \textit{catastrophic forgetting}—the loss of previously acquired knowledge when adapting to new data. The objective is to enable adaptation to $\mathcal{D}_T$ while preserving performance on $\mathcal{D}_S$, using minimal parameters. 

To address this, the model first learns from $\mathcal{D}_S$, after which separate branches are maintained for the source and each target distribution $\mathcal{D}_T$. Selective fine-tuning is applied only to the target branches, preserving the source branch to retain \textit{original performance}. Additionally, \textit{parameter-efficient adaptation} is achieved by updating only a subset of parameters while sharing the rest, enabling near-optimal target performance with no forgetting and reduced memory overhead (Fig.~\ref{fig:method}(b)). During inference, a supervised contrastive model classifies the input distribution and routes it to the corresponding branch for prediction (Sec.~\ref{scl}). Fig.~\ref{fig:method} illustrates the CLRerNet~\cite{honda2024clrernet} architecture in (a) and an example of our routing framework in (b) and (c).


\subsection{Modular Fine-Tuning for Forgetting Analysis}

In this study, we analyze which components of $f_\theta$ are most effective for adaptation and most susceptible to catastrophic forgetting. To this end, we perform hierarchical fine-tuning of subsets of $\theta$, evaluating their impact on performance across both $\mathcal{D}_S$ and $\mathcal{D}_T$. We define a set of fine-tuning configurations $\mathcal{C}$, where each $c \in \mathcal{C}$ updates a specific subset of model parameters $\theta'_c \subseteq \theta$, while the remaining modules are frozen. The resulting model is expressed as:

\begin{equation}
f_{\theta'_c}(x) = \mathcal{H}_{\theta_H'} \left( \mathcal{N}_{\theta_N'} \left( \mathcal{B}_{\theta_B'}(x) \right) \right)
\label{eq:modular_finetune}
\end{equation}

where $\theta'_c = \{\theta_B', \theta_N', \theta_H'\}$ specifies which components are trainable during adaptation. We evaluate the following configurations:

\begin{itemize}
    \item \textbf{Bias-only:} $\theta'_c = \{\text{bias}(\theta_i)\}$, where $i \in \{H\}$ or $\{N, H\}$.
    \item \textbf{Head-only:} $\theta'_c = \{\theta_H\}$.
    \item \textbf{Neck + Head:} $\theta'_c = \{\theta_N, \theta_H\}$.
    \item \textbf{Partial Backbone:} $\theta'_c = \{\theta_B^{(k)}, \theta_N, \theta_H\}$, where $\theta_B^{(k)}$ denotes the final $k$ layers of the backbone.
    \item \textbf{Full Fine-Tuning:} $\theta'_c = \{\theta_B, \theta_N, \theta_H\}$.
\end{itemize}

We provide detailed motivations for each configuration in Supplementary Material Section~\ref{motivation}.

\begin{figure*}[htbp] 
    \centering
    \includegraphics[width=0.8\linewidth]{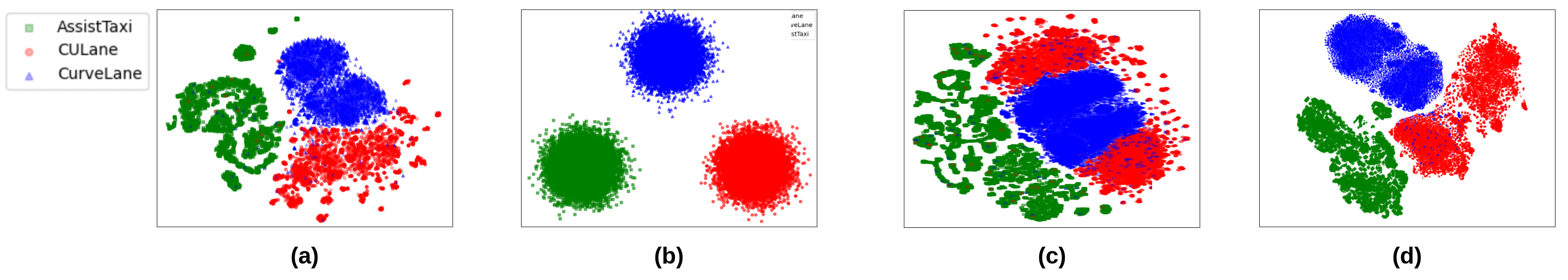}  
    \caption{
t-SNE visualization of distribution embeddings before and after contrastive learning. \textbf{(a)} Train embeddings before contrastive learning, showing limited separation. \textbf{(b)} Train embeddings after contrastive learning, with improved clustering by distribution. \textbf{(c)} Test embeddings before contrastive learning, showing poor separation. \textbf{(d)} Test embeddings after contrastive learning, where the projection head enables strong inter-distribution separation, supporting accurate centroid-based classification.
}
    \label{fig:my_figure}
\end{figure*}

\subsection{Distribution Classifier with Supervised
Constrastive Learning.}
\label{scl}
Initially, to characterize the nature of distribution shift, we examined how road surfaces from CULane~\cite{pan2018spatial}, CurveLanes~\cite{xu2020curvelane}, and AssistTaxi~\cite{ganeriwala2023assisttaxi} are organized in the embedding space of an ImageNet-pretrained ResNet encoder $\text{enc}_\phi(\cdot)$. The encoder extracts features from road surface regions that encode structured visual patterns---such as lane edge orientation, texture gradients, and spatial arrangement of road elements---and projects them into an embedding space where samples from the three datasets form loosely separated, distribution-specific clusters. To sharpen these boundaries and make them actionable, we subsequently trained a supervised contrastive learning (SCL) model that projects the encoder features into a space with greater inter-distribution separation, as illustrated in Figure~\ref{fig:my_figure}. The SCL model enables centroid-based classification for assignment of each input to its corresponding distribution during inference, thereby supporting routing to the dataset-specific branch of the lane detection model (Figure~\ref{fig:method}).

Given an input image $x_i \in \bigcup_{k=1}^{K} \mathcal{D}_k$ with distribution label $d_i \in \{1, \dots, K\}$, we extract features using the ResNet encoder $\text{enc}_\phi(\cdot)$ followed by an MLP projection $g_\psi(\cdot)$:

\begin{equation}
h_i = \text{enc}_\phi(x_i), \quad z_i = g_\psi(h_i).
\end{equation}

The model is trained end-to-end using a supervised contrastive loss~\cite{khosla2020supervised} over a minibatch $\mathcal{I}$:
\begin{equation}
\mathcal{L}_{\text{SupCon}} = \sum_{i \in \mathcal{I}} \frac{-1}{|P(i)|} \sum_{p \in P(i)} \log \frac{\exp(\mathrm{sim}(z_i, z_p)/\tau)}{\sum_{a \in \mathcal{A}(i)} \exp(\mathrm{sim}(z_i, z_a)/\tau)},
\end{equation}
where $P(i)$ is the set of positives sharing the same distribution label as $i$, $\mathcal{A}(i)$ includes all other samples in the batch, $\mathrm{sim}(\cdot, \cdot)$ denotes cosine similarity, and $\tau$ is a temperature parameter. After training, we compute a centroid for each distribution $\mathcal{D}_k$ as:
\begin{equation}
\mu_k = \frac{1}{|\mathcal{D}_k|} \sum_{i \in \mathcal{D}_k} z_i, \quad \text{for } k = 1, \dots, K.
\end{equation}

Subsequently, we classify the input $x$ by assigning it to the closest centroid in the embedding space:
\begin{equation}
\hat{d}(x) = \arg\min_{k \in {1, \dots, K}} ; |g_\psi(\text{enc}_\phi(x)) - \mu_k|_2^2.
\end{equation}

Given this predicted label $\hat{d}(x)$, we route the input through its corresponding model branch. For each distribution $\mathcal{D}_k$, we define a trainable parameter subset $\theta_k' \subseteq \theta$ represented by:
\begin{equation}
f_{\theta_k'}(x) = \mathcal{H}_{\theta_H^{(k)}} \left( \mathcal{N}_{\theta_N^{(k)}} \left( \mathcal{B}_{\theta_B^{(k)}}(x) \right) \right),
\end{equation}
where the modules may be shared or distribution-specific depending on the branching option. The final output is therefore:
\begin{equation}
\hat{y} = f_{\theta_{\hat{d}(x)}'}(x).
\end{equation}

\section{Experimental Results}
\label{sec:experiments}

\subsection{Experimental Setup}

\textbf{Datasets.}  
We evaluate our approach on three diverse datasets that exhibit significant variation in distribution, scene layout, and lane geometry. CULane~\cite{pan2018spatial} is a large-scale urban lane detection dataset with 88k training images, 9.7k validation images, and 34.7k test images. It includes scene-specific tags such as \textit{Normal}, \textit{Crowded}, and \textit{Curve}, capturing a wide range of traffic scenarios. CurveLanes~\cite{liu2021condlanenet} contains over 100k images with diverse and curved lane topologies. Following prior work~\cite{liu2021condlanenet}, we use the 20k validation split for evaluation. AssistTaxi is a challenging real-world dataset of airport taxiways with limited labels. For lane detection experiments, we use the available 5.5k labeled images: 3.5k for training and 2k for testing. For contrastive learning, we leverage all 85k images, partitioned into 65k for training and 20k for testing. The CULane and CurveLanes splits used for contrastive learning align with their standard train/test protocols.

\begin{figure}[h]
    \centering
    \includegraphics[width=\linewidth]{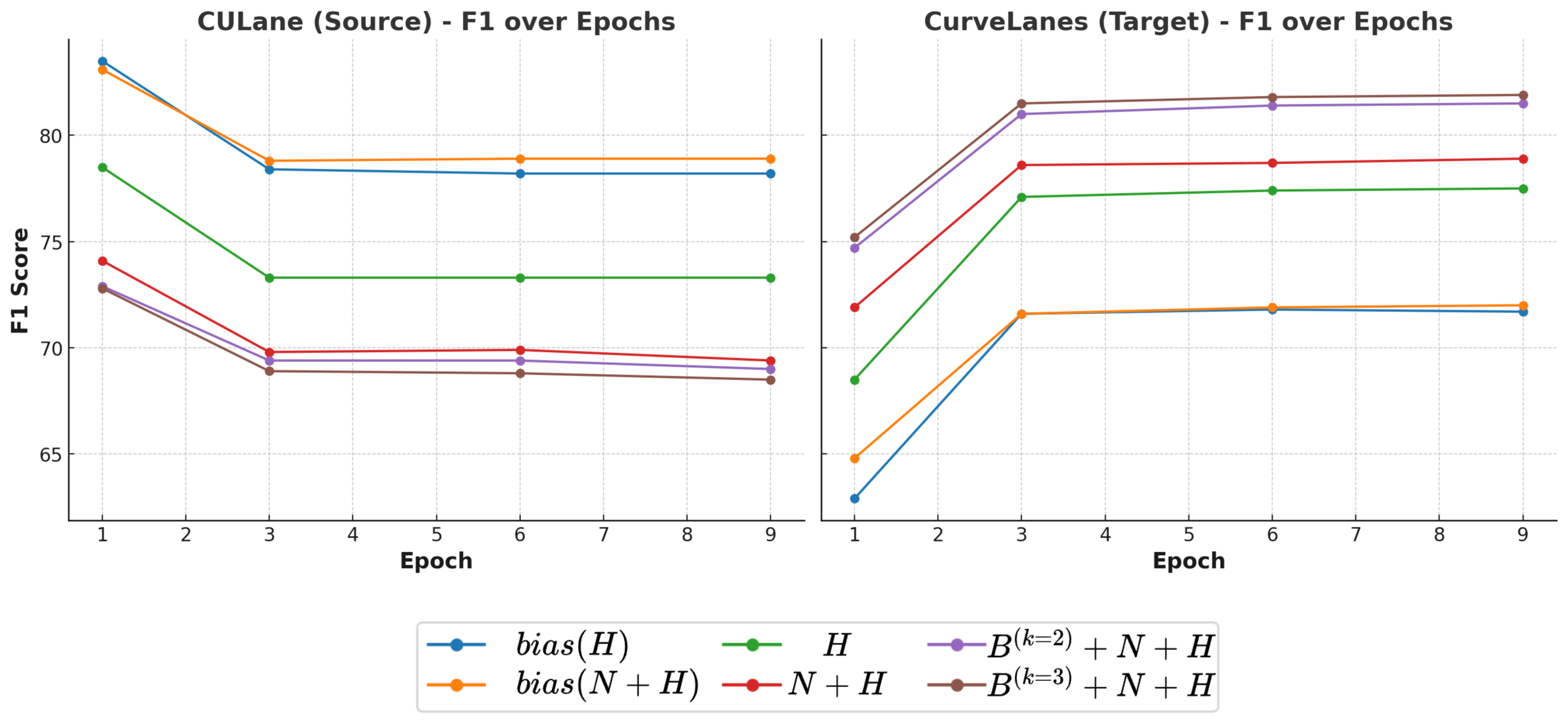}
    \caption{
    F1-scores plateau after three fine-tuning epochs, with modest gains on CurveLanes (target distribution) and little to no decline on CULane (source distribution) thereafter.
    }
    \label{fig:epoch_ablation}
\end{figure}

All experiments are conducted using the CLRerNet~\cite{honda2024clrernet} architecture with encoder backbones including DLA-34, ResNet-18, and ERFNet. Models are implemented in PyTorch and MMDetection~\cite{chen2019mmdetection}. Each model is first trained on a source distribution (CULane, CurveLanes, or AssistTaxi) for 15 epochs using the Adam optimizer with an initial learning rate of $6 \times 10^{-4}$ and cosine decay schedule. We adopt the data augmentation strategy from~\cite{honda2024clrernet}. We evaluate fine-tuning configurations involving the head (\texttt{H}), neck (\texttt{N}), and backbone (\texttt{B}), including variants with bias-only updates and partial backbone tuning. All experiments are run for three epochs with a fixed learning rate of $6 \times 10^{-4}$, based on trends in Figure~\ref{fig:epoch_ablation} which indicate that target performance saturates beyond this point.

\subsection{Evaluation Metrics}

We report F1 score as the primary metric, following the evaluation protocol in CLRerNet~\cite{honda2024clrernet}. We also report Precision and Recall in Supplementary pdf, for Table~\ref{tab:f1_drops_gains_params}. Specifically, F1 is computed via linear assignment between predicted and ground-truth lanes using an IoU matrix derived from rendered masks with a 30-pixel lane width and a matching threshold of $\text{tIoU} = 0.5$.

\begin{table*}[t]
\centering
\footnotesize
\setlength{\tabcolsep}{5pt}
\caption{F1-scores (at a confidence threshold of 0.5). The base model is trained on CULane, and results are shown after fine-tuning different components of model on CurveLanes and AssistTaxi. For each setting, we report the CULane F1-score with the corresponding \textit{drop} from its original performance, the target distribution F1-score with the \textit{gain} over the zero-shot baseline, and the number of trainable parameters.}
\label{tab:f1_drops_gains_params}
\begin{tabular}{ll|cc|cc|r}
\toprule
\textbf{Backbone} & \textbf{FT Config} &
\multicolumn{2}{c|}{\textbf{After fine-tuning with CurveLanes}} &
\multicolumn{2}{c|}{\textbf{After fine-tuning with AssistTaxi}} &
\textbf{\#Params} \\
\cmidrule(lr){3-4} \cmidrule(lr){5-6}
& & CULane: F1 (drop) & CurveLanes: F1 (gain) & CULane: F1 (drop) & AssistTaxi: F1 (gain) & \\
\midrule
\multirow{8}{*}{DLA-34}
& No fine-tuning & 81.2 & 65.0 & 81.2 & 0.0 & 0 \\
& bias(H) & 78.4 (-2.8) & 71.6 (+6.6) & 19.3 (-61.9) & 17.4 (+17.4) & 1,330 \\
& bias(N+H) & 78.8 (-2.4) & 71.6 (+6.6) & 20.5 (-60.7) & 18.7 (+18.7) & 1,714 \\
& H & 73.3 (-7.9) & 77.1 (+12.1) & 2.0 (-79.2) & 77.8 (+77.8) & 432,450 \\
& N+H & 70.8 (-10.4) & 78.6 (+13.6) & 0.6 (-80.6) & 82.1 (+82.1) & 600,770 \\
& B$^{(k=2)}$+N+H & 70.2 (-11.0) & 81.0 (+16.0) & 16.6 (-64.6) & 93.7 (+93.7) & 14,475,794 \\
& B$^{(k=3)}$+N+H & 68.5 (-12.7) & 81.5 (+16.5) & 2.8 (-78.4) & 92.5 (+92.5) & 15,682,834 \\
& B+N+H & 68.7 (-12.5) & 81.5 (+16.5) & 3.1 (-78.1) & 94.8 (+94.8) & 15,829,874 \\
\midrule
\multirow{8}{*}{ResNet-18}
& No fine-tuning & 80.4 & 64.4 & 80.4 & 0.0 & 0 \\
& bias(H) & 78.6 (-1.8) & 71.0 (+6.6) & 19.2 (-61.2) & 21.2 (+21.2) & 1,315 \\
& bias(N+H) & 78.8 (-1.6) & 70.6 (+6.2) & 16.2 (-64.2) & 19.0 (+19.0) & 1,699 \\
& H & 74.0 (-6.4) & 76.4 (+12.0) & 5.1 (-75.3) & 76.9 (+76.9) & 429,555 \\
& N+H & 72.1 (-8.3) & 77.8 (+13.4) & 5.4 (-75.0) & 82.8 (+82.8) & 597,875 \\
& B$^{(k=2)}$+N+H & 70.4 (-10.0) & 79.7 (+15.3) & 13.1 (-67.3) & 90.6 (+90.6) & 8,991,603 \\
& B$^{(k=3)}$+N+H & 70.0 (-10.4) & 79.8 (+15.4) & 16.3 (-64.1) & 93.5 (+93.5) & 11,091,315 \\
& B+N+H & 68.7 (-11.7) & 80.0 (+15.6) & 16.3 (-64.1) & 94.5 (+94.5) & 11,774,387 \\
\midrule
\multirow{8}{*}{ERFNet}
& No fine-tuning & 79.1 & 65.9& 79.1 & 0.0 & 0 \\
& bias(H) & 76.4 (-2.7) & 71.0 (+5.1) & 14.6 (-64.5) & 1.2 (+1.2) & 1,315 \\
& bias(N+H) & 76.3 (-2.8) & 71.2 (+5.3) & 17.5 (-61.6) & 1.7 (+1.7) & 1,699 \\
& H & 71.2 (-7.9) & 75.2 (+9.3) & 0.8 (-78.3) & 80.1 (+80.1) & 429,555 \\
& N+H & 68.9 (-10.2) & 76.6 (+10.7) & 0.8 (-78.3) & 86.6 (+86.6) & 597,875 \\
& B$^{(k=2)}$+N+H & 67.5 (-11.6) & 79.1 (+13.2) & 11.3 (-67.8) & 92.7 (+92.7) & 5,277,215 \\
& B$^{(k=3)}$+N+H & 66.9 (-12.2) & 79.3 (+13.4) & 6.1 (-73.0) & 90.9 (+90.9) & 5,474,847 \\
& B+N+H & 66.5 (-12.6) & 79.0 (+13.1) & 3.7 (-75.4) & 92.8 (+92.8) & 5,561,695 \\
\bottomrule
\end{tabular}
\end{table*}

\subsection{Quantitative Results}

\subsubsection*{Fine-tuning on Target Distributions Improves Target F1 Score but Reduces Source F1 Score.}

In Table~\ref{tab:f1_drops_gains_params}, we analyze the performance when CULane is used as the source and CurveLanes and AssistTaxi as target distributions.



When fine-tuning only the \texttt{bias(H)} and \texttt{bias(N+H)} parameters, the model adjusts activation thresholds---i.e., the confidence levels at which anchor priors trigger predictions---without altering feature representations or spatial priors. These lightweight updates (\(<2K\) parameters) improves performance on CurveLanes, which shares anchor priors and side-lane structures with CULane, allowing effective recalibration of anchor confidences and row-wise offsets (target F1: 71.0, source F1: 78.6). In contrast, AssistTaxi’s centralized taxiway structure demands large threshold shifts to activate center-aligned anchors and suppress side anchors, raising AssistTaxi F1 to 21.2 but severely disrupting source-specific anchor activations, collapsing CULane F1 score to 19.2. (ResNet-18 results discussed in this section; similar trends were observed with DLA-34 and ERFNet, as shown in Table~\ref{tab:f1_drops_gains_params}.)


Finetuning the head (\texttt{H}) offers more capacity (\textasciitilde430K parameters) and significantly improves target alignment (CurveLanes: 76.4, AssistTaxi: 76.9). The head samples anchor-aligned features and refines them via cross-attention with the global map, followed by fully connected layers that regress anchor parameters \((x_a, y_a, \theta_a, l)\), row-wise offsets, and confidence scores. This allows the model to capture CurveLanes’ lane curvature and topologies (F1: 76.4) and adapt to AssistTaxi’s centralized lane-layout (F1: 76.9), demonstrating the head’s capacity to reparameterize anchor priors under shift. However, with a frozen backbone and neck, the head operates on fixed source-distribution features, limiting its ability to reinterpret the target distribution well. After fine-tuning, the anchor priors---sampled from frozen, source-aligned features---can become misaligned under distribution shift, leading to degraded decoding fidelity and severe forgetting. This is evident in the AssistTaxi setting, where target F1 reaches 76.9, but source F1 collapses to 5.1.



Adding the neck module (\texttt{N+H}) substantially improves adaptation by enabling recalibration of intermediate feature representations. The neck redistributes spatial attention and refines resolution-specific semantics to better align with target lane geometries. Combined with the head, this configuration supports both upstream feature shaping and downstream anchor refinement. With only \textasciitilde600K parameters (5.1\% of total), \texttt{N+H} achieves strong target performance: F1 of 77.8 on CurveLanes and 82.8 on AssistTaxi with moderate source degradation (72.1 on ResNet-18). This suggests the frozen backbone retains transferable structure, while the adapted neck enables spatial realignment. 



\begin{table*}[htbp]
\centering
\small
\setlength{\tabcolsep}{4pt}
\caption{Comparing F1, Precision, Recall with AssistTaxi as the source, CULane and CurveLanes as target datasets, and a confidence threshold of 0.5. The table shows comparison of adaptation gains on targets and corresponding source degradation (leftmost column).}

\label{tab:assistaxi_transfer}
\begin{tabular}{ll|ccc|ccc|ccc}
\toprule
\footnotesize
\textbf{Backbone} & \textbf{FT Config} &
\multicolumn{3}{c|}{\textbf{AssistTaxi (Src)}} &
\multicolumn{3}{c|}{\textbf{CULane (Tgt)}} &
\multicolumn{3}{c}{\textbf{CurveLanes (Tgt)}} \\
& & F1 & Prec & Rec & F1 & Prec & Rec & F1 & Prec & Rec \\
\midrule
\multirow{6}{*}{DLA-34}
& No fine-tuning & 94.3 & 96.9 & 91.9 & 0.0 & 0.0 & 0.0 & 0.0 & 1.7 & 0.0 \\
& bias(H) & 0.0 & 0.0 & 0.0 & 0.0 & 0.0 & 0.0 & 0.0 & 0.0 & 0.0 \\
& bias(N+H) & 0.0 & 0.0 & 0.0 & 0.0 & 0.0 & 0.0 & 0.0 & 0.0 & 0.0 \\
& H & 0.0 & 0.0 & 0.0 & 3.8 & 52.9 & 2.0 & 1.1 & 6.8 & 0.6 \\
& N+H & 0.0 & 0.0 & 0.0 & 33.7 & 70.4 & 22.2 & 8.6 & 30.0 & 5.0 \\
& B$^{(k=2)}$ + N+H & 0.0 & 0.0 & 0.0 & 70.3 & 76.8 & 61.5 & 70.9 & 85.8 & 60.4 \\
& B$^{(k=3)}$ + N+H & 0.0 & 0.0 & 0.0 & 73.8 & 79.2 & 69.0 & 73.1 & 88.6 & 62.2 \\
& B+N+H & 0.0 & 0.0 & 0.0 & 76.3 & 81.3 & 71.9 & 76.5 & 91.4 & 65.8 \\
\bottomrule
\end{tabular}
\end{table*}

\begin{figure*}[htbp]
    \centering
    \includegraphics[width=0.8\linewidth]{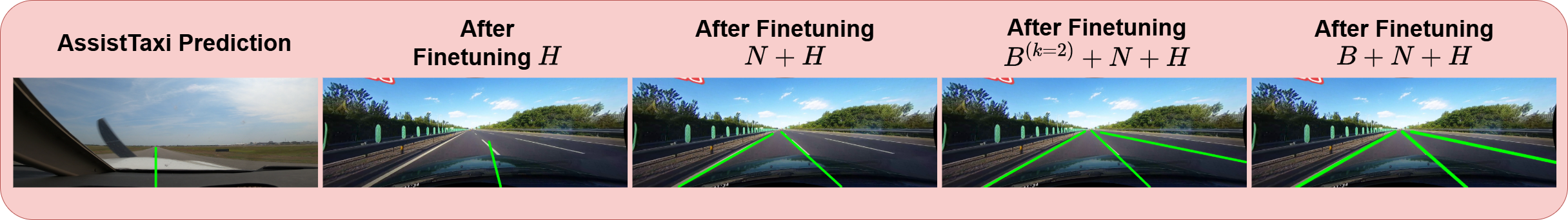}
    \caption{
Comparing AssistTaxi $\rightarrow$ CULane. 
\texttt{H:} Fine-tuning the head adjusts anchor regressors, detecting the center lane but failing on sides due to fixed upstream features. 
\texttt{N+H:} Adapting the neck improves detection for left and center lanes, but right lane remains undetected due to limited receptive adaptation.
\texttt{B$^{(k=2)}$+N+H}, \texttt{B+N+H:} Enable all lanes detection through deeper representation alignment.
}
    \label{fig:qualitative}
\end{figure*}

Extending the adaptation to deep backbone (B) layers (\texttt{B$^{(k=2)}$+N+H}, where $k$ denotes the last $k$ layers of B) enhances the model’s capacity to reshape mid-level features---such as lane textures and road boundaries---which feed into the FPN and modulate the neck’s multi-scale aggregation. While this adds over 11M parameters (18$\times$ more than \texttt{N+H}), it yields only modest gains on CurveLanes (F1: 79.8 vs. 77.8), suggesting that head and neck tuning suffice for this shift. In contrast, for AssistTaxi---where the distribution diverges from CULane---partial backbone adaptation proves critical, raising F1 from 82.8 to 93.5. This reflects the need to re-encode spatial cues specific to centralized taxiways that the neck alone cannot capture when operating on frozen source features. Interestingly, partial backbone finetuning on AssistTaxi also improves source retention: \texttt{B$^{(k=3)}$+N+H} recovers source F1 to 16.3 (vs. 5.1 for \texttt{H} and 5.4 for \texttt{N+H}). This behavior may seem counterintuitive, but it reflects the benefit of allowing selective adaptation in the deeper backbone layers. When only \texttt{H} or \texttt{N+H} is finetuned, these modules must compensate for fixed, potentially misaligned features which leads to overfitting and severe forgetting. In contrast, adapting the deeper backbone allows the model to recalibrate mid-level representations to the target distribution while preserving low-level source-aligned features, thereby reducing catastrophic forgetting.


\texttt{B+N+H} yields the best target performance (CurveLanes: 80.0, AssistTaxi: 94.5) but causes severe forgetting, particularly in source recall (see Supplementary pdf), indicating that new lane layouts are learned at the cost of overwriting prior knowledge. While longer training (15 epochs) using all parameters would likely restore baseline performance of a model, all results are reported after only three epochs to ensure fair comparison of adaptation efficiency. Similar trends were observed when using CurveLanes as the source, with progressive adaptation gains and forgetting patterns on CULane and AssistTaxi, consistent with Table~\ref{tab:f1_drops_gains_params}.

\begin{table*}[t]
\centering
\footnotesize
\caption{
Supervised Contrastive Learning (SCL)-based routing under distribution-specific specialization, using the \textbf{ERFNet} backbone. Each block corresponds to a source distribution, with F1\textsubscript{Src} constant across configurations (denoted with `` ''). The configuration shown in Figure~\ref{fig:method}(b.1) and (c.1), using CULane as the source and fine-tuning the \texttt{N+H}, is highlighted in red. Blue highlights indicate the most effective routing point for CurveLanes, and green for AssistTaxi (corresponding to Figure~\ref{fig:method}(b.2) and (c.2)), where Routing@\texttt{B$^{(k=2)}$+N+H} is optimal. The “Relative” column reports parameter efficiency relative to Routing@\texttt{B+N+H}, which is treated as 100\%.
}
\label{tab:routing}
\begin{tabular}{llcccccc}
\toprule
\footnotesize
\textbf{Source} & \textbf{Branch Point} & \textbf{F1\textsubscript{Src}} & \textbf{F1\textsubscript{Curve}} & \textbf{F1\textsubscript{Assist}} & \textbf{F1\textsubscript{Avg}} & \textbf{Params (M)} & \textbf{Relative} \\
\midrule
\multirow{4}{*}{CULane}
& Routing@\texttt{B+N+H}        & 79.1 & 79.0 & 92.8 & 83.6 & 16.7 & -- \\
& Routing@\texttt{B$^{(k=2)}$+N+H}     & `` ''  & 79.1 & 92.7 & 83.6 & 16.1 & 96\%  \\
& \cellcolor{red!20}Routing@\texttt{N+H}            & \cellcolor{red!20}`` ''   & \cellcolor{red!20}76.6 & \cellcolor{red!20}86.6 & \cellcolor{red!20}80.8 & \cellcolor{red!20}6.8  & \cellcolor{red!20}41\%  \\
& Routing@\texttt{H}                 & `` ''   & 75.2 & 80.1 & 78.1 & 6.4  & 38\%  \\
\midrule
\multicolumn{8}{c}{\textit{Next block uses CurveLanes as source, targets = CULane and AssistTaxi}} \\
\midrule
\textbf{Source} & \textbf{Branch Point} & \textbf{F1\textsubscript{Src}} & \textbf{F1\textsubscript{CULane}} & \textbf{F1\textsubscript{Assist}} & \textbf{F1\textsubscript{Avg}} & \textbf{Params (M)} & \textbf{Relative} \\
\midrule
\multirow{4}{*}{CurveLanes}
& Routing@\texttt{B+N+H}        & 84.8 & 78.6 & 91.7 & 85.0 & 16.7 & -- \\
& Routing@\texttt{B$^{(k=2)}$+N+H}     & `` ''   & 78.5 & 90.4 & 84.6 & 16.1 & 96\%  \\
& \cellcolor{blue!20}Routing@\texttt{N+H}            & \cellcolor{blue!20}`` ''   & \cellcolor{blue!20}75.7 & \cellcolor{blue!20}84.5 & \cellcolor{blue!20}81.7 & \cellcolor{blue!20}6.8  & \cellcolor{blue!20}41\%  \\
& Routing@\texttt{H}                 & `` ''   & 73.8 & 78.3 & 79.0 & 6.4  & 38\%  \\  
\midrule
\multicolumn{8}{c}{\textit{Next block uses AssistTaxi as source, targets = CULane and Curvelanes}} \\
\midrule
\textbf{Source} & \textbf{Branch Point} & \textbf{F1\textsubscript{Src}} & \textbf{F1\textsubscript{CULane}} & \textbf{F1\textsubscript{Curve}} & \textbf{F1\textsubscript{Avg}} & \textbf{Params (M)} & \textbf{Relative} \\
\midrule
\multirow{4}{*}{AssistTaxi}
& Routing@\texttt{B+N+H}         & 94.3 & 77.2 & 75.3 & 82.3 & 16.7 & -- \\
& \cellcolor{green!20}Routing@\texttt{B$^{(k=2)}$+N+H}     & \cellcolor{green!20}`` ''   & \cellcolor{green!20}71.8 & \cellcolor{green!20}71.5 & \cellcolor{green!20}79.2 & \cellcolor{green!20}16.1 & \cellcolor{green!20}96\%  \\
& Routing@\texttt{N+H}            & `` ''   & 30.4 & 6.0  & 43.6 & 6.8  & 41\%  \\
& Routing@\texttt{H}                 & `` ''   & 2.2  & 1.1  & 32.5 & 6.4  & 38\%  \\
\bottomrule
\end{tabular}
\end{table*}

In Table~\ref{tab:assistaxi_transfer}, when trained on AssistTaxi and evaluated on CULane or CurveLanes without fine-tuning, the model fails completely, yielding zero F1 across both target distributions. This stems from the dataset nature of AssistTaxi lanes: a single central taxiway line marking. The anchor priors and regressors are biased toward this centralized layout, making them fundamentally misaligned with the side-lane appearance of CULane and CurveLanes. \texttt{bias(H)} and \texttt{bias(N+H)} lack the capacity to adjust spatial priors or reweight features, and thus fail to activate the appropriate anchors for the target. These updates shift activation thresholds without changing feature representations, causing the model to lose its original decision boundaries and collapse source performance to zero.


Starting from \texttt{H}, we observe initial adaptation gains on the target distributions, most notably CULane (F1: 3.8), with further improvements as the neck is included and deeper backbone layers are progressively unfrozen (shown in Figure~\ref{fig:qualitative}). The detection head alone updates anchor-specific regressors and classifiers, enabling partial detection of lane structures in CULane and Curvelanes. However, with a fixed neck and backbone, the model continues to operate on AssistTaxi-specific features, limiting adaptation to decoding-level adjustments. Incorporating the neck (\texttt{N+H}) recalibrates multi-scale feature fusion, boosting target F1 (CULane: 33.7, CurveLanes: 8.6), though representation capacity remains constrained by the frozen backbone. As more of the backbone is adapted (\texttt{B$^{(k=2)}$+N+H}, \texttt{B$^{(k=3)}$+N+H}), mid-level features reorient toward side-lane structures, pushing target F1 to 70–73. Full model fine-tuning (\texttt{B+N+H}) yields the highest target performance (CULane: 76.3, CurveLanes: 76.5). Across all configurations, however, the AssistTaxi F1 remains at zero, indicating complete forgetting of the original centralized anchor priors. We observed similar trends on ResNet-18 and ERFNet backbones, consistent with the DLA-34 results in Table~\ref{tab:assistaxi_transfer}.

\subsection{Inference-Time Contrastive Routing}

The distribution classifier using supervised contrastive learning (SCL) model achieves near-perfect accuracy across all three datasets—\textbf{99.6\%} on CULane, \textbf{99.7\%} on CurveLanes, and \textbf{99.9\%} on AssistTaxi. “Routing@\texttt{X}” indicates the point where distribution-specific branches begin: layers before \texttt{X} are shared, while those from \texttt{X} onward are cloned and fine-tuned for each target. At inference, the distribution classifier routes the input to the dataset-specific branch. Using the lightest backbone, ERFNet, full inference takes an average of \textbf{9.82 ms} per image on an NVIDIA RTX 4080. Contrastive routing adds only \textbf{0.31 ms}, resulting in just \textbf{3.2\%} overhead while maintaining real-time feasibility.

\subsubsection*{Target Branching Enables Adaptation Without Sacrificing Source Performance}
In Table~\ref{tab:routing}, each block corresponds to a model pretrained on one source distribution and adapted to others via modular branching, which benefits from additional parameters via fine-tuned branches; this means cloning and fine-tuning only specific modules (\texttt{H}, \texttt{N+H}, and \texttt{B$^{(k=2)}$+N+H}) for the target distribution, while keeping the original branch fixed to preserve source performance, as illustrated in Figure~\ref{fig:method}.

ERFNet is a lightweight backbone replacing standard residual blocks with non-bottleneck 1D modules~\cite{romera2017erfnet}. In Table~\ref{tab:routing}, we show results using the ERFNet backbone and observe high parameter efficiency across routing configurations. Compared to fine-tuning fully separate models (treated as baseline), routing at partial backbone branches achieves high target performance (Avg F1: 83.6, 84.6, 79.2) at 96\% of the parameters. When the source-target shift is moderate (CULane$\rightarrow$CurveLanes/AssistTaxi), Routing@\texttt{N+H} strikes the best balance, achieving near-optimal performance with only 41\% of the parameters. For more severe shifts (AssistTaxi$\rightarrow$CULane/Curvelane), deeper Routing@\texttt{B$^{(k=2)}$+N+H} remains essential, as lighter branches (\texttt{H}, \texttt{N+H}) fail to adapt effectively. 


\section{Limitations}

While we demonstrate our method on CLRerNet, it can be similarly applied to other lane detection architectures, and we encourage future work to explore this generalization. Additionally, although our routing framework mitigates forgetting during adaptation, it requires maintaining separate branches for each distribution, leading to increased parameter overhead as the number of target distributions grows. We implore the community to explore student–teacher distillation or shared-feature decoders as future directions to consolidate knowledge across branches and further improve generalization and parameter efficiency under shift.

\section{Conclusion}

In this work, we presented a component-wise analysis of fine-tuning strategies for lane detection under distribution shift, revealing the trade-offs between adaptation and catastrophic forgetting. We proposed a contrastive routing framework that dynamically routes inputs to dataset-specific branches, enabling strong target generalization while preserving source performance using fewer parameters than training separate end-to-end models. We ran evaluations across multiple backbones, including ERFNet, showing its suitability for distribution-aware deployment.

\section{Acknowledgements}

This work was supported by NASA Langley Research Center (LaRC) under Award No. 80LARC23DA003. We thank Dr. Philip K. Chan for his guidance in deepening our understanding of supervised contrastive loss and for his valuable feedback during the writing process, which improved the clarity and presentation of this paper.

{
    \small
    \bibliographystyle{ieeenat_fullname}
    \bibliography{main}
}

\clearpage
\setcounter{page}{1}
\maketitlesupplementary

\FloatBarrier

\section{Rationale for Fine-Tuning Configurations}
\label{motivation}
Here, we provide the motivations for each configuration $\theta'_c \subseteq \theta$, outlining the intuition behind the choice of trainable components and their impact under distribution shift.

\begin{itemize}
    \item \textbf{Bias:} $\theta'_c = \{\text{bias}(\theta_i)\}$, where $i \in \{H\}$ or $\{N, H\}$. Updating only the bias terms allows localized adaptation with minimal parameter updates, offering a lightweight alternative to mitigate distribution shift while largely preserving source representations.
    
    \item \textbf{Head:} $\theta'_c = \{\theta_H\}$. Fine-tuning the detection head refines anchor-specific predictions and confidence calibration to better align with the LaneIoU metric. 

    \item \textbf{Neck + Head:} $\theta'_c = \{\theta_N, \theta_H\}$. Adding the neck enables adaptation of feature aggregation across scales, allowing the model to better represent lane geometries that vary in curvature, spacing, or resolution in the target distribution.

    \item \textbf{Partial Backbone:} $\theta'_c = \{\theta_B^{(k)}, \theta_N, \theta_H\}$, where $\theta_B^{(k)}$ denotes the final $k$ layers of the backbone. This configuration enables adaptation of higher-level features sensitive to lane geometry, curvature, and anchor offsets, while preserving early low-level features to minimize forgetting.

    \item \textbf{Full Fine-Tuning:} $\theta'_c = \{\theta_B, \theta_N, \theta_H\}$. This setting provides full adaptation capacity but risks erasing generalizable or source-specific knowledge, especially in lower-level feature extractors.
\end{itemize}

\section{Additional Results}
\label{sec:rationale}

We present detailed precision, recall, and F1-score breakdowns for all fine-tuning configurations in Tables~\ref{tab:curve_full_metrics_deltas} and~\ref{tab:assist_full_metrics_deltas}, corresponding to CurveLanes and AssistTaxi target distributions, respectively. Each table reports both source (CULane) and target performance after selectively fine-tuning model components. In addition to F1-score, which is analyzed in the main paper, these tables reveal trends in precision and recall that offer deeper insight into the nature of distribution adaptation and forgetting. For every metric, we include the relative \textit{drop} from the no-finetuning CULane baseline and the \textit{gain} over the zero-shot baseline on the target distribution, highlighting trade-offs between forgetting and adaptation across configurations.

\begin{table*}[!t]
\centering
\small
\setlength{\tabcolsep}{5pt}
\caption{The base model is trained on CULane, and results are shown after fine-tuning different components on CurveLanes. Precision, Recall, and F1-score are reported at a confidence threshold of 0.5. For each setting, we show source (CULane) and target (CurveLanes) performance, with relative \textit{drops} from the CULane no-finetuning baseline and \textit{gains} over the CurveLanes zero-shot baseline indicated in parentheses.}
\label{tab:curve_full_metrics_deltas}
\begin{tabular}{ll|ccc|ccc|r}
\toprule
\textbf{Backbone} & \textbf{FT Config} &
\multicolumn{3}{c|}{\textbf{CULane (Src)}} &
\multicolumn{3}{c|}{\textbf{CurveLanes (Tgt)}} &
\textbf{\#Params} \\
\cmidrule(lr){3-5} \cmidrule(lr){6-8}
& & F1 (drop) & Prec (drop) & Rec (drop) & F1 (gain) & Prec (gain) & Rec (gain) & \\
\midrule
\multirow{8}{*}{DLA-34}
& No fine-tuning & 81.2 & 89.3 & 74.4 & 65.0 & 94.4 & 49.5 & 0 \\
& bias(H) & 78.4 (-2.8) & 81.4 (-7.9) & 75.6 (+1.2) & 71.6 (+6.6) & 89.4 (-5.0) & 59.7 (+10.2) & 1,330 \\
& bias(N+H) & 78.8 (-2.4) & 82.9 (-6.4) & 75.2 (+0.8) & 71.6 (+6.6) & 89.1 (-5.3) & 59.9 (+10.4) & 1,714 \\
& H & 73.3 (-7.9) & 73.0 (-16.3) & 73.6 (-0.8) & 77.1 (+12.1) & 90.8 (-3.6) & 67.1 (+17.6) & 432,450 \\
& N+H & 70.8 (-10.4) & 69.3 (-20.0) & 72.3 (-2.1) & 78.6 (+13.6) & 91.7 (-2.7) & 68.8 (+19.3) & 600,770 \\
& B$^{(k=2)}$+N+H & 70.2 (-11.0) & 69.0 (-20.3) & 71.2 (-3.2) & 81.0 (+16.0) & 93.6 (-0.8) & 71.4 (+21.9) & 14,475,794 \\
& B$^{(k=3)}$+N+H & 68.5 (-12.7) & 68.1 (-21.2) & 69.0 (-5.4) & 81.5 (+16.5) & 94.1 (-0.3) & 71.9 (+22.4) & 15,682,834 \\
& B+N+H & 68.7 (-12.5) & 68.2 (-21.1) & 69.1 (-5.3) & 81.5 (+16.5) & 94.0 (-0.4) & 72.0 (+22.5) & 15,829,874 \\
\midrule
\multirow{8}{*}{ResNet-18}
& No fine-tuning & 80.4 & 87.9 & 74.0 & 64.4 & 92.9 & 49.3 & 0 \\
& bias(H) & 78.6 (-1.8) & 82.7 (-5.2) & 74.9 (+0.9) & 71.0 (+6.6) & 89.5 (-3.4) & 58.9 (+9.6) & 1,315 \\
& bias(N+H) & 78.8 (-1.6) & 83.8 (-4.1) & 74.4 (+0.4) & 70.6 (+6.2) & 89.1 (-3.8) & 58.5 (+9.2) & 1,699 \\
& H & 74.0 (-6.4) & 75.4 (-12.5) & 72.5 (-1.5) & 76.4 (+12.0) & 90.6 (-2.3) & 66.1 (+16.8) & 429,555 \\
& N+H & 72.1 (-8.3) & 72.8 (-15.1) & 71.5 (-2.5) & 77.8 (+13.4) & 91.8 (-1.1) & 67.4 (+18.1) & 597,875 \\
& B$^{(k=2)}$+N+H & 70.4 (-10.0) & 71.5 (-16.4) & 69.4 (-4.6) & 79.7 (+15.3) & 93.5 (+0.6) & 69.5 (+20.2) & 8,991,603 \\
& B$^{(k=3)}$+N+H & 70.0 (-10.4) & 71.9 (-16.0) & 68.2 (-5.8) & 79.8 (+15.4) & 94.1 (+1.2) & 69.3 (+20.0) & 11,091,315 \\
& B+N+H & 68.7 (-11.7) & 69.8 (-18.1) & 67.7 (-6.3) & 80.0 (+15.6) & 94.2 (+1.3) & 69.6 (+20.3) & 11,774,387 \\
\midrule
\multirow{8}{*}{ERFNet}
& No fine-tuning & 79.1 & 85.6 & 73.6 & 65.9 & 92.0 & 51.4 & 0 \\
& bias(H) & 76.4 (-2.7) & 78.6 (-7.0) & 74.2 (+0.6) & 71.0 (+5.1) & 87.4 (-4.6) & 59.8 (+8.4) & 1,315 \\
& bias(N+H) & 76.3 (-2.8) & 78.3 (-7.3) & 74.4 (+0.8) & 71.2 (+5.3) & 87.3 (-4.7) & 60.1 (+8.7) & 1,699 \\
& H & 71.2 (-7.9) & 72.2 (-13.4) & 70.3 (-3.3) & 75.2 (+9.3) & 89.6 (-2.4) & 64.8 (+13.4) & 429,555 \\
& N+H & 68.9 (-10.2) & 69.3 (-16.3) & 68.6 (-5.0) & 76.6 (+10.7) & 90.4 (-1.6) & 66.4 (+15.0) & 597,875 \\
& B$^{(k=2)}$+N+H & 67.5 (-11.6) & 68.6 (-17.0) & 66.4 (-7.2) & 79.1 (+13.2) & 92.9 (+0.9) & 68.9 (+17.5) & 5,277,215 \\
& B$^{(k=3)}$+N+H & 66.9 (-12.2) & 67.4 (-18.2) & 66.5 (-7.1) & 79.3 (+13.4) & 92.7 (+0.7) & 69.3 (+17.9) & 5,474,847 \\
& B+N+H & 66.5 (-12.6) & 68.0 (-17.6) & 65.0 (-8.6) & 79.0 (+13.1) & 92.9 (+0.9) & 68.7 (+17.3) & 5,561,695 \\
\bottomrule
\end{tabular}
\end{table*}

\begin{table*}[t]
\centering
\small
\setlength{\tabcolsep}{5pt}
\caption{The base model is trained on CULane, and results are shown after fine-tuning different components on AssistTaxi. Precision, Recall, and F1-score are reported at a confidence threshold of 0.5. For each setting, we show source (CULane) and target (AssistTaxi) performance, with relative \textit{drops} from the CULane no-finetuning baseline and \textit{gains} over the AssistTaxi zero-shot baseline indicated in parentheses.}
\label{tab:assist_full_metrics_deltas}
\begin{tabular}{ll|ccc|ccc|r}
\toprule
\textbf{Backbone} & \textbf{FT Config} &
\multicolumn{3}{c|}{\textbf{CULane (Src)}} &
\multicolumn{3}{c|}{\textbf{AssistTaxi (Tgt)}} &
\textbf{\#Params} \\
\cmidrule(lr){3-5} \cmidrule(lr){6-8}
& & F1 (drop) & Prec (drop) & Rec (drop) & F1 (gain) & Prec (gain) & Rec (gain) & \\
\midrule
\multirow{8}{*}{DLA-34}
& No fine-tuning & 81.2 & 89.3 & 74.4 & 0.0 & 0.0 & 0.0 & 0 \\
& bias(H) & 19.3 (-61.9) & 48.6 (-40.7) & 12.1 (-62.3) & 17.4 (+17.4) & 82.9 (+82.9) & 9.7 (+9.7) & 1,330 \\
& bias(N+H) & 20.5 (-60.7) & 48.6 (-40.7) & 12.9 (-61.5) & 18.7 (+18.7) & 77.5 (+77.5) & 10.7 (+10.7) & 1,714 \\
& H & 2.0 (-79.2) & 2.4 (-86.9) & 1.7 (-72.7) & 77.8 (+77.8) & 80.5 (+80.5) & 75.4 (+75.4) & 432,450 \\
& N+H & 0.6 (-80.6) & 10.9 (-78.4) & 0.3 (-74.1) & 82.1 (+82.1) & 85.3 (+85.3) & 79.1 (+79.1) & 600,770 \\
& B$^{(k=2)}$+N+H & 16.6 (-64.6) & 83.8 (-5.5) & 9.2 (-65.2) & 93.7 (+93.7) & 96.3 (+96.3) & 91.3 (+91.3) & 14,475,794 \\
& B$^{(k=3)}$+N+H & 2.8 (-78.4) & 72.6 (-16.7) & 1.4 (-73.0) & 92.5 (+92.5) & 93.6 (+93.6) & 91.5 (+91.5) & 15,682,834 \\
& B+N+H & 3.1 (-78.1) & 75.5 (-13.8) & 0.7 (-73.7) & 94.8 (+94.8) & 97.3 (+97.3) & 92.5 (+92.5) & 15,829,874 \\
\midrule
\multirow{8}{*}{ResNet-18}
& No fine-tuning & 80.4 & 87.9 & 74.0 & 0.0 & 0.0 & 0.0 & 0 \\
& bias(H) & 19.2 (-61.2) & 74.2 (-13.7) & 11.0 (-63.0) & 21.2 (+21.2) & 70.9 (+70.9) & 12.5 (+12.5) & 1,315 \\
& bias(N+H) & 16.2 (-64.2) & 70.4 (-17.5) & 9.1 (-64.9) & 19.0 (+19.0) & 66.9 (+66.9) & 11.1 (+11.1) & 1,699 \\
& H & 5.1 (-75.3) & 61.2 (-26.7) & 2.6 (-71.4) & 76.9 (+76.9) & 79.7 (+79.7) & 74.3 (+74.3) & 429,555 \\
& N+H & 5.4 (-75.0) & 61.8 (-26.1) & 2.8 (-71.2) & 82.8 (+82.8) & 84.7 (+84.7) & 81.0 (+81.0) & 597,875 \\
& B$^{(k=2)}$+N+H & 13.1 (-67.3) & 85.6 (-2.3) & 7.1 (-66.9) & 90.6 (+90.6) & 91.9 (+91.9) & 89.3 (+89.3) & 8,991,603 \\
& B$^{(k=3)}$+N+H & 16.3 (-64.1) & 37.5 (-50.4) & 10.4 (-63.6) & 93.5 (+93.5) & 94.9 (+94.9) & 92.1 (+92.1) & 11,091,315 \\
& B+N+H & 16.3 (-64.1) & 85.3 (-2.6) & 9.0 (-65.0) & 94.5 (+94.5) & 97.2 (+97.2) & 91.9 (+91.9) & 11,774,387 \\
\midrule
\multirow{8}{*}{ERFNet}
& No fine-tuning & 79.1 & 85.6 & 73.6 & 0.0 & 0.0 & 0.0 & 0 \\
& bias(H) & 14.6 (-64.5) & 50.6 (-35.0) & 8.5 (-65.1) & 1.2 (+1.2) & 10.3 (+10.3) & 0.7 (+0.7) & 1,315 \\
& bias(N+H) & 17.5 (-61.6) & 54.7 (-30.9) & 10.4 (-63.2) & 1.7 (+1.7) & 25.7 (+25.7) & 1.0 (+1.0) & 1,699 \\
& H & 0.8 (-78.3) & 49.4 (-36.2) & 0.4 (-73.2) & 80.1 (+80.1) & 85.2 (+85.2) & 75.6 (+75.6) & 429,555 \\
& N+H & 0.8 (-78.3) & 34.6 (-51.0) & 0.4 (-73.2) & 86.6 (+86.6) & 88.3 (+88.3) & 85.0 (+85.0) & 597,875 \\
& B$^{(k=2)}$+N+H & 11.3 (-67.8) & 84.5 (-1.1) & 6.1 (-67.5) & 92.7 (+92.7) & 95.1 (+95.1) & 90.5 (+90.5) & 5,277,215 \\
& B$^{(k=3)}$+N+H & 6.1 (-73.0) & 85.7 (+0.1) & 3.2 (-70.4) & 90.9 (+90.9) & 92.3 (+92.3) & 89.6 (+89.6) & 5,474,847 \\
& B+N+H & 3.7 (-75.4) & 90.6 (+5.0) & 1.9 (-71.7) & 92.8 (+92.8) & 94.9 (+94.9) & 90.8 (+90.8) & 5,561,695 \\
\bottomrule
\end{tabular}
\end{table*}


\end{document}